\documentclass[conference]{IEEEtran}
\IEEEoverridecommandlockouts
\usepackage{cite}
\usepackage{amsmath,amssymb,amsfonts}
\usepackage{graphicx}
\usepackage{textcomp}
\usepackage{xcolor}
    \usepackage{xcolor} 
    \usepackage{hyperref}
    \usepackage{graphicx}
    \usepackage{array}
    \usepackage{listings}
\usepackage{algorithm2e}
\usepackage{algpseudocode}
\usepackage{booktabs}
    \usepackage{amsmath} 
\def\BibTeX{{\rm B\kern-.05em{\sc i\kern-.025em b}\kern-.08em
    T\kern-.1667em\lower.7ex\hbox{E}\kern-.125emX}}
\begin{document}

    \title{ERPA: Efficient RPA Model Integrating OCR and LLMs for Intelligent Document Processing
}

\author{\IEEEauthorblockN{Osama hosam abdellaif}
\IEEEauthorblockA{\textit{Computer Science} \\
\textit{MSA university}\\
Cairo, Egypt \\
osama.hosam@msa.edu.eg}
\and
\IEEEauthorblockN{Abdelrahman Nader Hassan}
\IEEEauthorblockA{\textit{Computer Science} \\
\textit{MSA university}\\
Cairo, Egypt \\
abdelrahman.nader@msa.edu.eg}
\and
\IEEEauthorblockN{Ali Hamdi}
\IEEEauthorblockA{\textit{Computer Science} \\
\textit{MSA university}\\
Cairo, Egypt \\
ahamdi@msa.edu.eg}
}

\IEEEoverridecommandlockouts
\IEEEpubid{\makebox[\columnwidth]{ 979-8-3503-6777-5/24/\$31.00 ©2024 IEEE\hfill}
\hspace{\columnsep}\makebox[\columnwidth]{ }}
\maketitle
\IEEEpubidadjcol

\begin{abstract}
This paper presents ERPA, an innovative Robotic Process Automation (RPA) model designed to enhance ID data extraction and optimize Optical Character Recognition (OCR) tasks within immigration workflows. Traditional RPA solutions often face performance limitations when processing large volumes of documents, leading to inefficiencies. ERPA addresses these challenges by incorporating Large Language Models (LLMs) to improve the accuracy and clarity of extracted text, effectively handling ambiguous characters and complex structures. Benchmark comparisons with leading platforms like UiPath and Automation Anywhere demonstrate that ERPA significantly reduces processing times by up to 94\%, completing ID data extraction in just 9.94 seconds. These findings highlight ERPA's potential to revolutionize document automation, offering a faster and more reliable alternative to current RPA solutions.
\end{abstract}


\section{Introduction}

In the evolving global landscape of immigration, the demand for high-performance, scalable, and precise document processing has intensified. With the exponential growth in the volume immigration document like IDs , passports, visas, and certificates—manual data extraction and verification face critical challenges in terms of efficiency, accuracy, and throughput. Robotic Process Automation (RPA) has emerged as a potent solution for automating repetitive workflows, reducing human error, and optimizing operational efficiency \cite{c1, c2}. RPA automates tasks by mimicking human interactions with digital systems, handling activities such as data extraction, file manipulation, and automated report generation. These systems have become a key enabler of digital transformation across various sectors \cite{c3, c9}. A critical advancement in RPA has been the integration of Optical Character Recognition (OCR), which enhances automation by extracting text from scanned documents, improving the speed and accuracy of data processing \cite{c17,al2018arabic,c18, c19,al2018enhanced}. Despite these advancements, traditional RPA systems encounter significant challenges when dealing with unstructured or semi-structured documents.

Existing RPA platforms, such as UiPath and Automation Anywhere, are optimized for rule-based, structured data processing. However, they face performance bottlenecks when applied to unstructured or semi-structured data, particularly text extracted from scanned images or documents with complex formats \cite{c3, c9}. OCR systems often struggle with accuracy when processing documents containing ambiguous characters, dynamic layouts, or handwritten text \cite{hamdi2021c}. Moreover, scaling these systems to process large volumes of documents leads to inefficiencies in processing speed and error rates, which are critical in high-volume environments like immigration services \cite{c17, c18}. These limitations necessitate a more sophisticated approach that can handle complex document structures, maintain high accuracy, and improve processing efficiency. 

To address these challenges, we propose ERPA, an enhanced RPA framework that integrates advanced OCR technology with Large Language Models (LLMs)\cite{hassle2024} to enhance both the accuracy and adaptability of text extraction in document processing workflows. ERPA utilizes state-of-the-art OCR systems to extract textual content from diverse document formats and applies LLMs to refine and validate the extracted data, ensuring improved accuracy when dealing with ambiguous or complex structures. Additionally, ERPA dynamically adapts to varying document layouts, enabling seamless processing of immigration documents such as IDs ,passports, visas, and certificates in real-time \cite{c20, c23}. The proposed model introduces several technical innovations, including a multi-stage pipeline for document ingestion, OCR-based text extraction, and LLM-driven data interpretation and validation.

\begin{figure*}[h]\
    \centering\includegraphics[width=0.6\textwidth]{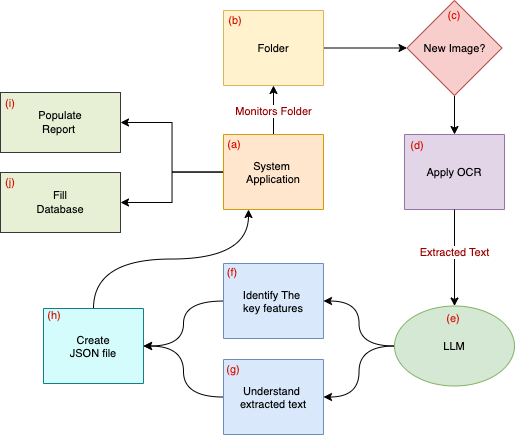} 
        \caption{ERPA system architecture: The diagram illustrates the workflow of the proposed ERPA model for automating ID document processing. The process begins with the system application monitoring the folder for new files (steps \textbf{a}, \textbf{b}). When a new image is detected (\textbf{c}), the system applies OCR to extract the relevant text features (\textbf{d}). The extracted text is then processed by a Large Language Model (LLM) (\textbf{e}), which helps to understand and interpret the text (\textbf{g}) while identifying key features (\textbf{f}). Once the text is interpreted, the system creates a structured JSON file (\textbf{h}) to help populating both the database (\textbf{j}) and a report (\textbf{i}). This architecture ensures efficient and accurate ID document processing while maintaining scalability for high-volume workflows.}

    \label{fqmodel}
\end{figure*}

As shown in Figure \ref{fqmodel}, the ERPA system architecture consists of multiple interconnected modules designed to optimize both the speed and accuracy of document processing. The system begins by ingesting scanned immigration documents, which are then passed through OCR for text extraction. The extracted data is subsequently processed by a fine-tuned LLM, which interprets the text, resolves ambiguities, and ensures that critical data fields—such as names, birthdates, and ID numbers—are correctly identified and validated. This multi-stage approach not only improves processing accuracy but also significantly reduces the time required to process each document. In our benchmarking tests, ERPA achieves a 93\% reduction in processing time compared to existing RPA platforms, completing ID data extraction in just 9.94 seconds per document \cite{c21}.

The technical capabilities of ERPA extend beyond basic text extraction and validation. By leveraging large language models (LLMs), ERPA adapts dynamically to new document formats without requiring manual configuration or predefined rules. This adaptability makes it highly suitable for processing the diverse and dynamic document structures encountered in immigration workflows. Additionally, the system is designed to scale efficiently, maintaining high throughput even when processing large volumes of documents. Through rigorous evaluation, we demonstrate that ERPA not only improves text extraction accuracy but also reduces overall processing time by 59\% compared to traditional OCR-RPA systems. This positions ERPA as a transformative solution for automating high-volume, document-heavy workflows, such as those seen in immigration services.

The remainder of this paper is organized as follows: Section \ref{rw} explores related work in RPA, OCR, and LLM integration; Section \ref{m} presents and analyzes the experimental results; Section \ref{rd} discusses the findings and contributions of the ERPA model; and Section \ref{c} concludes with the implications of our model for large-scale immigration document processing.

\section{Related Work}\label{rw}
In the context of immigration document processing, Robotic Process Automation (RPA) plays a critical role in automating the extraction and handling of data from various ID forms and records. Leading RPA tools such as UiPath and Automation Anywhere have been widely adopted to streamline repetitive tasks, including Optical Character Recognition (OCR)-based document handling \cite{c1, c2}. RPA simplifies labor-intensive tasks by mimicking human actions, enhancing the efficiency of industries that manage large volumes of transactions. These platforms can automate a range of tasks, from front-office duties to end-to-end processes, enabling employees to focus on more strategic activities \cite{c9}. RPA operates in three primary modes: attended, unattended, and hybrid. In attended mode, bots act as digital assistants, performing tasks such as logging into systems and retrieving data after receiving human commands. However, they require the operator to pause other system activities while the bot is running. Unattended bots operate independently, executing tasks remotely without human intervention, which is ideal for large-scale automation. Hybrid RPA combines both modes, enabling complex workflows where human decision-making is required before bots complete the automated tasks \cite{c10}.

Despite the effectiveness of these tools in automating repetitive tasks, significant limitations arise when handling unstructured data, often found in complex immigration documents. UiPath, for example, integrates well with OCR technologies like Google OCR and Microsoft OCR, but its performance declines when processing unstructured or ambiguous text \cite{c17}. Similarly, Automation Anywhere excels in combining RPA with analytics for data extraction, yet its script-heavy interface and limited flexibility for unstructured text pose challenges in real-world applications \cite{c18}. Both platforms are valuable, but their performance bottlenecks in OCR tasks highlight the need for more advanced solutions capable of handling the complexities of immigration documents.

Recent developments in Large Language Models (LLMs) have significantly enhanced multiple learning tasks \cite{hamad2024asem}. Specifically, LLMs have improved the capability of OCR systems, particularly when dealing with unstructured or poorly formatted text. By incorporating LLMs, OCR workflows can now interpret and extract more accurate information from documents with complex layouts, diverse fonts, or ambiguous characters \cite{c14}. LLMs allow automation to move beyond rigid rule-based systems, enabling greater adaptability to the varied and unpredictable nature of immigration forms. Despite these advancements, there are still gaps in reproducibility and transparency in the studies comparing different RPA tools and methodologies. Many works fail to provide accessible datasets or publicly available code, limiting the ability to replicate results and benchmark performance. Furthermore, comparative analyses that include Python-based custom implementations versus commercial RPA solutions are lacking, particularly in terms of efficiency and scalability \cite{c20}. Addressing these research gaps is vital for advancing the field of OCR-driven RPA, especially in high-stakes environments such as immigration systems where accuracy and speed are paramount.

\section{ERPA Model Research Methodology}\label{m}

\begin{figure*}[h]
\centering    \includegraphics[width=0.8\textwidth]{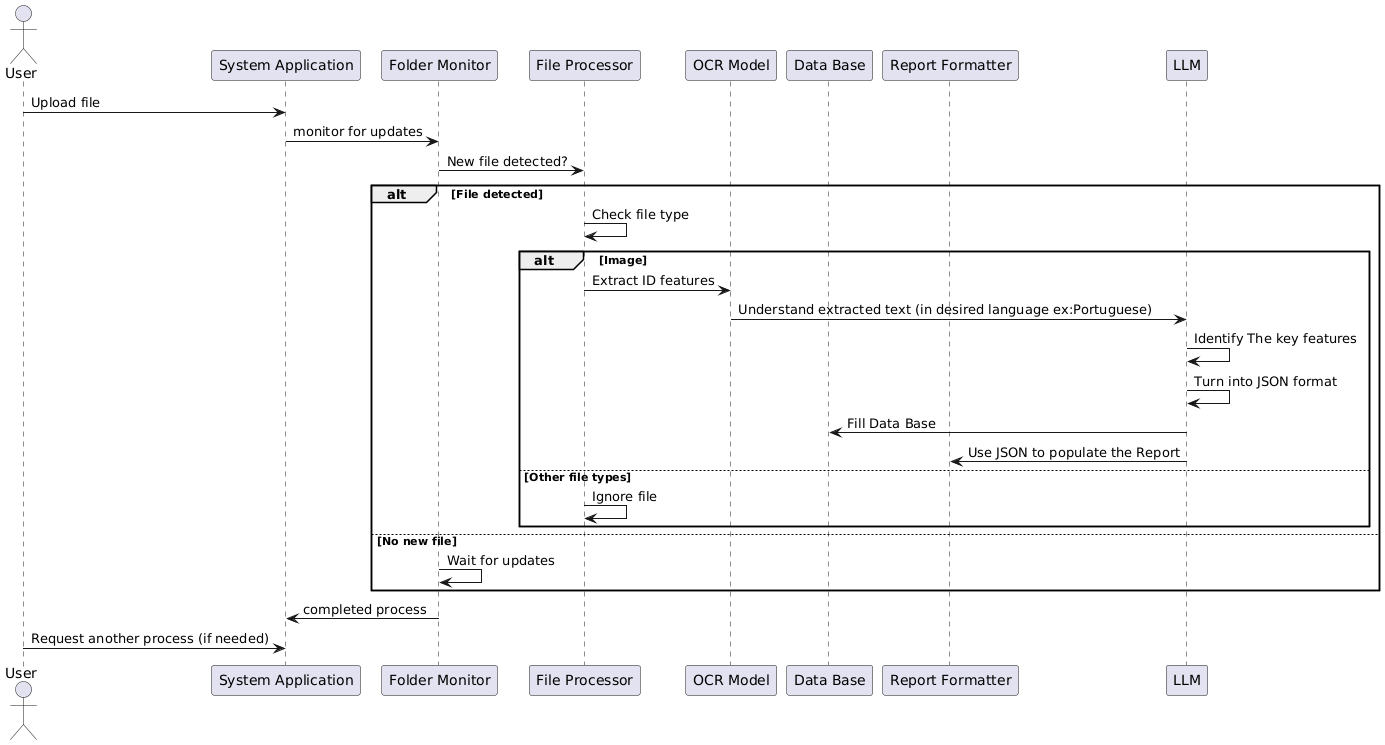} 
    \caption{ERPA system architecture: The diagram illustrates the workflow of the proposed ERPA model for automating ID document processing. The process begins with the system application monitoring the folder for new files (steps \textbf{a}, \textbf{b}). When a new image is detected (\textbf{c}), the system applies OCR to extract the relevant text features (\textbf{d}). The extracted text is then processed by a Large Language Model (LLM) (\textbf{e}), which helps to understand and interpret the text (\textbf{g}) while identifying key features (\textbf{f}). Once the text is interpreted, the system creates a structured JSON file (\textbf{h}) and populates both the database (\textbf{j}) and a report (\textbf{i}). This architecture ensures efficient and accurate ID document processing while maintaining scalability for high-volume workflows.}
    \label{seqmodel}
\end{figure*}

We present the ERPA model, a fully automated system designed to streamline the extraction and processing of data from immigration documents. This model continuously monitors a designated directory for new files, such as passports, birth certificates, and visa documents, as shown in Figure \ref{seqmodel}. When a valid file is detected, ERPA utilizes advanced Optical Character Recognition (OCR) engines to extract the relevant text, which is subsequently processed by a Large Language Model (LLM) to generate structured JSON outputs. The organized data is automatically stored in a database, where it can be used to generate formatted reports, thus improving the accuracy and efficiency of handling diverse document formats \cite{c10, c14}. ERPA is specifically designed to address the complexities of processing immigration-related documents.

\subsection{System Architecture Overview}

The ERPA system is designed to handle various document formats and types with high efficiency. The core system architecture comprises several stages: directory monitoring, OCR-based text extraction, LLM-based data structuring, and automated report generation. Each component is integrated to ensure seamless processing of immigration documents.

\subsection{Directory Monitoring and File Detection}

The system initiates by continuously monitoring a designated directory, denoted as $\mathcal{D}$, for any new files. This monitoring process operates in an infinite loop, periodically checking for changes within the directory. When a new image file $f_i$ is detected within the valid file set $\mathcal{I}$, it triggers the processing sequence. The detection mechanism compares the set of files at time $t$ with those at time $t - \Delta t$, represented as:
\begin{equation}
    f_i \in \mathcal{F}(t) \setminus \mathcal{F}(t - \Delta t)
\end{equation}

This method ensures that the system quickly detects and responds to new files.

\subsection{Text Extraction via OCR}

Upon detecting a valid image file, the system proceeds with text extraction using one of two Optical Character Recognition (OCR) engines: PaddleOCR\cite{c20} or DocTR\cite{c24}. The choice of OCR engine is based on the document's format and language. The extracted text data is represented as $ \mathbf{T}$ and computed as:
\begin{equation}
    \mathbf{T} = \mathcal{O}_k(f_i), \quad k \in \{1, 2\}
\end{equation}
This step ensures that the textual content from the image is accurately captured for further processing, regardless of the document's complexity or language.

\subsection{Data Structuring via LLM}

After OCR extraction, the raw text $ \mathbf{T}$ is processed by a Large Language Model (LLM) $\mathcal{L}$, which interprets the content, identifies key features such as names, dates, and ID numbers, and structures the data in a predefined JSON format $ \mathbf{J}$:
\begin{equation}
\mathbf{J} = \mathcal{L}(\mathbf{T})
\end{equation}

This structured data format facilitates easier storage, retrieval, and manipulation. The LLM is fine-tuned to handle variations in document formats and languages, ensuring robustness across diverse immigration documents.

\subsection{Data Storage and Report Generation}

Once the data has been structured into JSON format, it is automatically stored in a database for further use. The data is also mapped into an Excel sheet $ \mathcal{E}$ through a defined mapping function $\mathcal{M}$:
\begin{equation}
\mathcal{E} = \mathcal{M}(\mathbf{J})
\end{equation}

This structured Excel data forms the foundation for generating formatted reports in Word document format $ \mathcal{W}$ using the function:
\begin{equation}
\mathcal{W} = \mathcal{F}_{\text{doc}}(\mathcal{E})
\end{equation}

This automated process ensures that all necessary data is accurately reflected in the reports, thus improving overall efficiency in document management and reporting.

\subsection{Continuous Automation and Scalability}

The entire ERPA workflow operates in a continuous loop, detecting, processing, and generating reports for each new file added to the monitored directory. The system is designed to be highly scalable, capable of handling large volumes of data without compromising accuracy or speed. This continuous automation process is represented by the algorithm in \textbf{Algorithm 1}, ensuring that the ERPA system can operate efficiently in high-volume environments such as immigration offices.

\begin{algorithm}
\caption{ERPA Algorithm for ID Data Extraction}\label{alg:ERPA}
\KwData{Directory $\mathcal{D}$ containing immigration documents}
\KwResult{Structured data and formatted reports}
Initialize monitoring of the directory $\mathcal{D}$\;
\While{True}{
    Check for new files in $\mathcal{D}$\;
    \If{new file $f_i$ is detected}{
        \If{$f_i \in \mathcal{I}$ \Comment*[r]{(i.e., $f_i$ is an image)}}{
            Extract text data $\mathbf{T}$ using OCR engine $\mathcal{O}_k(f_i)$\;
            Process extracted text $\mathbf{T}$ using LLM $\mathcal{L}$\;
            Transform $\mathbf{T}$ into structured JSON $\mathbf{J}$\;
            Save structured JSON $\mathbf{J}$ to the database\;
            Populate Excel sheet $\mathcal{E}$ from structured data $\mathbf{J}$\;
            Generate formatted report $\mathcal{W}$ from Excel data $\mathcal{E}$\;
        }
        \Else{
            Ignore non-image file\;
        }
    }
    Continue monitoring\;
}
\end{algorithm}

\section{Benchmark and Experimental Design}

The benchmarking of the ERPA model against established Robotic Process Automation (RPA) tools, specifically UiPath and Automation Anywhere, was conducted using a rigorous experimental design. The primary objective of this benchmark was to evaluate the efficiency of each system in processing immigration-related documents, particularly focusing on Optical Character Recognition (OCR) processing speed. Two OCR engines—PaddleOCR and DocTR—were used across all models to ensure a comprehensive evaluation of processing efficiency.

\subsection{Dataset and Experimental Setup}

For the experiments, we curated a dataset consisting of 500 high-resolution images of Brazilian ID documents, sourced from the publicly available BID dataset. The images included variations in format, layout, and lighting conditions to simulate real-world document processing challenges. The dataset was preprocessed to ensure consistency in image resolution and quality before being fed into the models.

To ensure a fair comparison, each RPA tool ( UiPath, and Automation Anywhere)\cite{baweja_comparative} and our ERPA was configured to operate under identical conditions. All models were deployed on the same machine.

\subsection{Evaluation Criteria}
The benchmarking focused on the processing speed as a key performance metric. Measured in seconds, this metric tracks the time taken by each RPA model to process an entire batch of documents through the OCR engines. Processing speed is critical in high-volume immigration workflows where delays can lead to significant bottlenecks.

\subsection{Experimental Procedure}

The experimental procedure was divided into two phases:
\begin{itemize}
    \item \textbf{Phase 1 - Baseline Processing:} In this phase, each model was tested individually using PaddleOCR. All 500 images were processed sequentially, and the time taken for each batch was recorded.
    \item \textbf{Phase 2 - Advanced Processing:} This phase involved repeating the experiments using DocTR as the OCR engine with RPA tools (UIPath and Automation anywhere) and our Proposed model ERPA. Similarly, the processing times were recorded for all models.
\end{itemize}

To ensure reproducibility, each experiment was run three times, and the average processing times were calculated. Additionally, we evaluated the system's ability to handle multiple language translations, focusing on Portuguese, the official language in Brazil, to assess LLM integration performance.

\section{Results and Discussion}\label{rd}

This section presents a comprehensive analysis of the ERPA model’s performance compared to leading RPA tools, UiPath and Automation Anywhere, along with a detailed examination of the document structure, processing challenges, and common patterns encountered. Additionally, we provide a step-by-step example of ERPA’s document processing approach, illustrating the effectiveness of the methodology.

\subsection{Document Structure and Processing Challenges}

Documents such as IDs, passports, and visas used in immigration contexts typically exhibit a structured layout with predefined fields (e.g., name, birthdate, ID number). However, these documents often present a variety of challenges, including variations in layout, font size, and image resolution. Such complexities necessitate advanced preprocessing techniques to ensure reliable OCR results across different document types and quality levels.

A notable challenge in processing these documents involves managing different languages and alphabets, especially where non-Latin characters or diacritical marks are present. ERPA’s methodology includes tailored preprocessing adjustments to normalize these variations, enabling accurate OCR recognition even for multilingual documents\cite{malathi2021}. Furthermore, issues such as document wear or faded printing require additional noise reduction techniques to ensure clean and legible OCR outputs.

\subsection{Step-by-Step Example of ERPA Document Processing}

To illustrate the methodology, we provide a step-by-step example of ERPA processing an ID document using PaddleOCR:

\begin{enumerate}
    \item \textbf{Directory Monitoring and File Detection:} The ERPA system initiates by continuously monitoring a designated directory for new files, such as passports, birth certificates, and visa documents. When a new file is detected, the processing sequence is triggered to ensure timely extraction.

    \item \textbf{Text Extraction via OCR:} After preprocessing, the system applies Optical Character Recognition (OCR) using engines such as PaddleOCR or DocTR, depending on the document's language and structure. This step captures all relevant text from the document accurately.

    \item \textbf{Data Structuring via LLM:} The extracted text is then processed by a Large Language Model (LLM), which interprets the text, identifies key features, and structures the data into a standardized JSON format for easy storage and retrieval.

    \item \textbf{Data Storage and Export:} Verified data is saved in a CSV file or database. ERPA standardizes fields across document types, ensuring consistency in downstream processes and facilitating integration with immigration systems.

    \item \textbf{Automated Report Generation:} The data is mapped into an Excel template and formatted into a report, which is then exported as a Word document. This feature enables the generation of accurate, formatted reports directly from the structured data.

\end{enumerate}

\subsection{Performance Comparison}

Table \ref{tab} compares the performance of our custom ERPA model against UiPath and Automation Anywhere across two OCR engines, PaddleOCR and DocTR.

\begin{table}[ht]
\centering
\caption{Comparison of Automation Models by Task}
\label{tab}
\begin{tabular}{lcc}
\toprule
\textbf{Manual process} & 160 sec \\
\midrule
\textbf{Model} & \textbf{PaddleOCR} & \textbf{DocTR} \\ 
\midrule
\textbf{UiPath} & 16.8 sec & 16.7 sec \\ 
\textbf{Automation Anywhere} & 18.52 sec & 18.65 sec \\ 
\textbf{ERPA (Ours)} & 9.94 sec & 10.16 sec \\ 
\bottomrule
\end{tabular}
\end{table}

\subsection{Productivity and Time Savings}

In addition to speed, ERPA provides significant productivity improvements over manual document processing, which typically takes 160 seconds per ID document. Using PaddleOCR, ERPA reduced this processing time to 9.94 seconds, yielding a time savings of approximately 93.78\%. With DocTR, the processing time was 10.16 seconds, still achieving a 93.65\% reduction. UiPath and Automation Anywhere also reduced processing time significantly but did not match ERPA's efficiency, with time savings of 89.56\% and 88.34\%, respectively. These results underscore ERPA's ability to dramatically enhance productivity in routine document processing tasks.

\subsection{Comparative Performance of ERPA}

The ERPA model consistently outperformed UiPath and Automation Anywhere, achieving a time savings of 59\% over UiPath and 53.67\% over Automation Anywhere when using PaddleOCR. With DocTR, ERPA maintained a performance lead despite the higher computational demand, completing tasks in less than two-thirds of the time taken by its competitors. These results emphasize ERPA’s suitability for high-volume, repetitive tasks, where operational efficiency is paramount.

\subsection{Impact of Automation on Productivity}

ERPA's performance advantage over both UiPath and Automation Anywhere highlights the transformative potential of custom RPA solutions. By reducing processing time by over 90\% compared to manual methods, ERPA significantly increases productivity while allowing human operators to focus on tasks that require higher cognitive skills. The study further illustrates that even the least efficient automation solution reduced processing time to under 6.21\% of manual processing time, showcasing the potential of advanced automation in reducing operational burdens across various industries.

Overall, ERPA offers substantial productivity gains in document processing, with the potential to transform industries reliant on high-volume data extraction tasks. The results validate ERPA’s design and configuration as an effective automation solution, particularly suited for time-sensitive and accuracy-critical contexts such as immigration services.

\section{Data and Code Availability}
The code repository for this project can be found at: \href{https://github.com/MAD-SAM22/LMRPA}{GitHub Repository for LMRPA}.

The dataset used in this study is available at: \href{https://www.kaggle.com/datasets/osamahosamabdellatif/high-quality-brazilian-ids}{Project Dataset}.

\section{Conclusion}\label{c}

In summary, the ERPA model demonstrated substantial performance improvements over both UiPath and Automation Anywhere, with notable time savings and high levels of accuracy in data extraction. These results highlight ERPA's capacity to streamline complex document processing workflows, particularly in time-sensitive and accuracy-critical domains like immigration services. By automating the extraction of essential information and seamlessly integrating this data into immigration systems, the proposed ERPA solution addresses major challenges in document handling, significantly reducing manual workload and enhancing efficiency.

The integration of LLM-based extraction within ERPA further ensures that data retrieval is precise and adaptable to evolving regulatory requirements. This adaptability, combined with ERPA’s automated report generation feature, provides a comprehensive solution that not only speeds up processing but also improves the consistency and reliability of documentation workflows.

\subsection{Future Work}

Future research could focus on enhancing ERPA’s accuracy through an ensemble approach, applying majority voting across multiple LLMs in combination with various OCR engines. By running several LLMs alongside different OCR models and selecting outputs based on majority consensus, this approach could yield even higher accuracy, particularly in handling complex or low-quality document images. Additionally, exploring this ensemble method could improve robustness across diverse document formats and languages, further expanding ERPA's applicability in diverse real-world settings.

In conclusion, the proposed ERPA model serves as a foundational step toward automated document processing for high-stakes environments. The suggested future enhancements offer promising avenues for further research, advancing the capabilities of RPA systems in achieving reliable, adaptable, and highly accurate data extraction.

\bibliographystyle{unsrt}
\bibliography{main}

\end{document}